%% file: egbib.tex
\documentclass[10pt,twocolumn,letterpaper]{article}

\usepackage{iccv}
\usepackage{times}
\usepackage{epsfig}
\usepackage{graphicx}
\usepackage{amsmath}
\usepackage{amssymb}
\usepackage{bm}
\usepackage{threeparttable}
\usepackage{multirow}
\usepackage{amssymb}
\usepackage{framed,enumitem} 
\usepackage{array}
\usepackage{url}            
\usepackage{booktabs}       
\usepackage{amsfonts}       
\usepackage{nicefrac}       
\usepackage{microtype}      
\usepackage{subcaption}
\usepackage{float}
\usepackage{caption,overpic,textpos}
\usepackage[table]{xcolor}
\usepackage[british, english, american]{babel}

\usepackage[pagebackref=true,breaklinks=true,letterpaper=true,colorlinks,bookmarks=false]{hyperref}

\iccvfinalcopy 

\def\iccvPaperID{****} 

\ificcvfinal\fi

\newlength\savewidth

\renewcommand{\paragraph}[1]{\vspace{1.25mm}\noindent\textbf{#1}}

\newcolumntype{x}[1]{>{\centering\arraybackslash}p{#1pt}}
\newcolumntype{y}[1]{>{\raggedright\arraybackslash}p{#1pt}}
\newcolumntype{z}[1]{>{\raggedleft\arraybackslash}p{#1pt}}

\newcommand{\app}{\raise.17ex\hbox{$\scriptstyle\sim$}}

\definecolor{deemph}{gray}{0.6}

\definecolor{baselinecolor}{gray}{.9}

\begin{document}

\title{LexLIP: Lexicon-Bottlenecked Language-Image Pre-Training for\\Large-Scale Image-Text Retrieval}

\author{Ziyang Luo$^{1}$\thanks{Work done during the internship at Microsoft.}, Pu Zhao$^{2}$, Can Xu$^{2}$, Xiubo Geng$^{2}$, Tao Shen$^{2}$, Chongyang Tao$^{2}$,\\Jing Ma$^{1}$, Qingwen Lin$^{2}$, Daxin Jiang$^{2}$\thanks{ Corresponding author}\\
  $^1$  {\normalsize Hong Kong Baptist University, Hong Kong SAR, China} \\
  $^2$ {\normalsize Microsoft Corporation} \\
    \texttt{\normalsize cszyluo@comp.hkbu.edu.hk, majing@hkbu.edu.hk}\\
  \texttt{\normalsize \{pu.zhao,caxu,xigeng,shentao,chongyang.tao,qlin,djiang\}@microsoft.com}}

\maketitle
\ificcvfinal\thispagestyle{empty}\fi

\begin{abstract}
   Image-text retrieval (ITR) is a task to retrieve the relevant images/texts, given the query from another modality. The conventional \textbf{dense retrieval paradigm} relies on encoding images and texts into dense representations using dual-stream encoders, however, it faces challenges with low retrieval speed in large-scale retrieval scenarios.
   In this work, we propose the \textbf{lexicon-weighting paradigm}, where sparse representations in vocabulary space are learned for images and texts to take advantage of the bag-of-words models and efficient inverted indexes, resulting in significantly reduced retrieval latency. 
   A crucial gap arises from the continuous nature of image data, and the requirement for a sparse vocabulary space representation. To bridge this gap, we introduce a novel pre-training framework, \textbf{Lexicon-Bottlenecked Language-Image Pre-Training (LexLIP)}, that learns importance-aware lexicon representations. This framework features lexicon-bottlenecked modules between the dual-stream encoders and weakened text decoders, allowing for constructing continuous bag-of-words bottlenecks to learn lexicon-importance distributions.
   Upon pre-training with same-scale data, our \textbf{LexLIP} achieves state-of-the-art performance on two benchmark ITR datasets, MSCOCO and Flickr30k. Furthermore, in large-scale retrieval scenarios, \textbf{LexLIP} outperforms CLIP with a $5.5\sim221.3\times$ faster retrieval speed and $13.2\sim48.8\times$ less index storage memory.
\end{abstract}


\input{1.Introduction.tex}
\input{2.RelatedWork.tex}
\input{3.Methodology.tex}
\input{4.Experiment.tex}
\input{5.Conclusion.tex}

{\small
\bibliographystyle{ieee_fullname}
\bibliography{egbib}
}

\end{document}

%% file: 1.introduction.tex
\section{Introduction}

\begin{figure}
     \centering
     \begin{subfigure}[b]{0.23\textwidth}
         \centering
         \includegraphics[width=3.5cm]{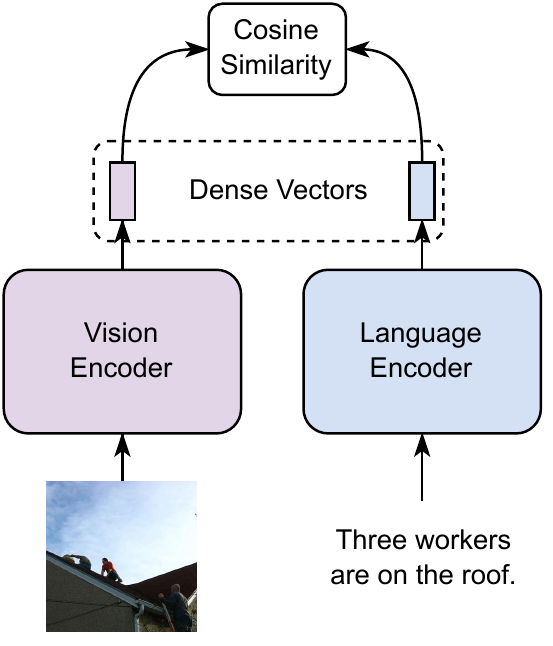}
         \caption{Dense Retrieval Paradigm.}
         \label{fig:dense}
     \end{subfigure}
     \begin{subfigure}[b]{0.24\textwidth}
         \centering
         \includegraphics[width=3.5cm]{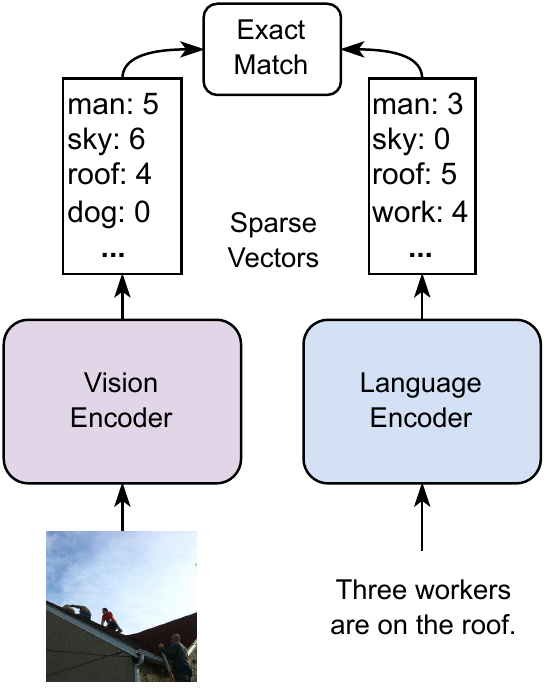}
         \caption{Lexicon-Weighting Paradigm.}
         \label{fig:sparse}
     \end{subfigure}
    \caption{Comparing the traditional dense retrieval paradigm and our brand-new lexicon-weighting paradigm.}
    \label{fig:intro}
\end{figure}

Image-text retrieval (ITR) is an important problem that involves retrieving relevant images based on textual queries and vice versa. It has numerous applications in different fields such as e-commerce product search~\cite{DBLP:conf/kdd/LiLJLYZWM21} and medical image retrieval~\cite{DBLP:conf/wacv/HuVH22}. Despite the importance of ITR, the majority of existing works~\cite{align,cots,clip,sun-etal-2021-lightningdot} have focused on small-scale retrieval tasks, such as MSCOCO~\cite{coco} and Flickr30k~\cite{f30k}. The number of images and texts to be retrieved is less than 30k. However, in real-world scenarios, such as Google Image Search, there is a growing need for large-scale ITR, where the number of samples exceeds 1M and the retrieval time becomes a critical concern. The current dense retrieval paradigm, in which each image and text is represented as a dense vector (as shown in Figure~\ref{fig:dense}), can become computationally expensive and result in slow retrieval speed. The large-scale k-nearest neighbor (KNN) dense retrieval requires calculating the cosine similarity between the query and all candidate samples, causing the retrieval time to increase linearly as the number of samples increases~\cite{lexmae}. This limitation makes the dense retrieval paradigm unsuitable for real-world applications and highlights the need for more efficient and effective methods for large-scale retrieval.

\begin{figure*}
    \centering
    \includegraphics[height=7cm]{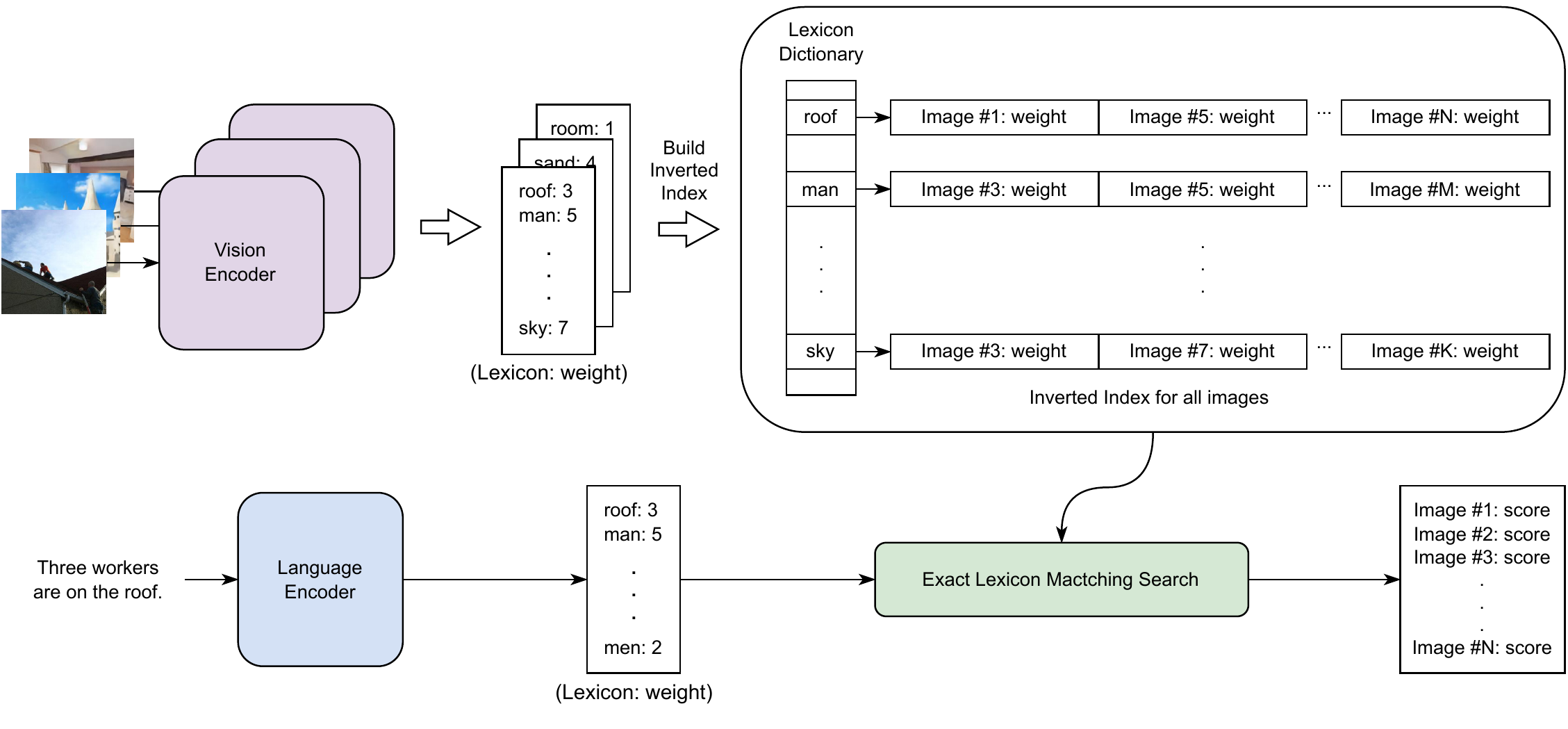}
    \caption{An overview of the Exact Lexicon Matching Search of our lexicon-weighting paradigm for text-to-image retrieval.}
    \label{fig:inverted}
\end{figure*}

In this work, we present a novel \textbf{lexicon-weighting paradigm} for ITR, as illustrated in Figure~\ref{fig:sparse}. This paradigm represents images and texts as sparse representations in the vocabulary space, where the relevant lexicons are assigned high weights and the other lexicons are set to 0. The retrieval process involves transforming these lexicon-weighted representations into inverted indexes, as depicted in Figure~\ref{fig:inverted}. Subsequently, the Exact Lexicon Matching Search algorithm~\cite{BM25} is employed to retrieve the relevant samples, which  only calculates similarity scores with candidate samples that contain matching lexicons, thereby eliminating the need to iterate over all samples and significantly reducing the retrieval latency. Additionally, this paradigm can make the best of lexicon-level contextualization by considering either high coordinate~\cite{splade} or concurrence~\cite{doc2query} visual objects and textual terms.

A crucial gap is that images are continuous data, whereas the lexicon-weighting paradigm requires projecting them into the sparse vocabulary space. To bridge this gap, we propose a novel pre-training framework, \textbf{Lexicon-Bottlenecked Language Image Pre-Training (LexLIP)}, to learn importance-aware lexicon representations. Thereby, \textbf{LexLIP} consists of three components: i) the dual-stream encoders, ii) two lexicon-bottlenecked modules, and iii) two weakened masking-style text decoders; and two pre-training phases: i) lexicon-bottlenecked pre-training; ii) momentum lexicon-contrastive pre-training.

Specifically, the first pre-training phase comprises four distinct objectives, namely image lexicon-bottlenecked masked language modeling, text lexicon-bottlenecked masked language modeling, text self-supervised masked language modeling, and in-batch lexicon contrastive learning. The first two objectives involve passing an image or masked text into a vision or language encoder, respectively, to produce token-level language modeling (LM) logits in the vocabulary space. A max-pooling operation followed by a normalization function is applied to the LM logits to derive a lexicon-importance distribution. This distribution is then utilized by the lexicon-bottlenecked module as weights to produce a continuous bag-of-words (CBoW) dense bottleneck. Meanwhile, the weakened masking-style text decoder is tasked with reconstructing the masked text from the bottleneck. Given the weakened decoder and aggressive masking, the decoder is inclined to recover masked tokens based on the CBoW bottleneck. As a result, the LexLIP encoders assign higher importance scores to crucial vocabulary lexicons of the image or text and lower scores to trivial ones, aligning closely with the objective of the lexicon-weighting retrieval paradigm and enhancing its performance.

The second phase pre-training in this work is the momentum lexicon-contrastive learning, where images and texts are aligned in the sparse vocabulary space with a large-scale negative sample size. Empirically, large negative sample sizes are crucial for good performance~\cite{cots,clip}. Therefore, we adopt the momentum contrastive learning of MoCo~\cite{moco} to cache negative samples, thereby decoupling the size of the negative samples from the mini-batch size. This approach employs two momentum encoders to update the samples in the queues, similar to previous studies~\cite{cots}.

The experimental results reveal that our \textbf{LexLIP} pre-trained with same-scale image-text pairs demonstrates state-of-the-art performance on the widely used ITR benchmarks, MSCOCO~\cite{coco} and Flickr30k~\cite{f30k}. In the large-scale retrieval scenario, \textbf{LexLIP} demonstrates a remarkable improvement in QPS with a 5.5 to 221.3 times increase and a significant reduction in index storage memory with a 13.2 to 48.8 times decrease, compared to CLIP~\cite{clip}.

Our contributions can be listed as follows:
\begin{enumerate}
    \item We are the first to introduce the \textbf{Lexicon-Weighting Paradigm} to ITR, which represents images and texts as sparse representations in the lexicon vocabulary space.
    \item We propose the novel \textbf{Lexicon-Bottlenecked Language Image Pre-Training (LexLIP)} to learn the lexicon-weighting representations.
    \item Our \textbf{LexLIP} achieves state-of-the-art performance on MSCOCO and Flickr30k, under fair comparison. In large-scale ITR, we achieve $5.5\sim221.3\times$ speed-up and $13.2\sim48.8\times$ less index storage memory than CLIP.
\end{enumerate}

%% file: 2.RelatedWork.tex
\input{figure_mae.tex}

\section{Related Work}

\textbf{Image-Text Retrieval.} ITR has received considerable attention in the cross-modal community. Traditional approaches to ITR utilized Convolutional Neural Networks (CNNs)~\cite{CNN} as encoders to individually encode images and texts~\cite{vse,DBLP:journals/pami/WangLHL19}. In recent years, the popularity of transformer-based models and large-scale language-image pre-training have seen a surge~\cite{align,cots,conlip,clip,sun-etal-2021-lightningdot}. These models have achieved state-of-the-art performance on various ITR benchmarks. However, the dense retrieval paradigm adopted by these models results in low retrieval speed, making them unsuitable for large-scale image-text retrieval applications. Our proposed \textbf{Lexicon-Bottlenecked Language Image Pre-Training (LexLIP)} method introduces a novel lexicon-weighting paradigm, resulting in significantly improved retrieval speed.

\input{figure_lexmoco.tex}

\textbf{Lexicon-Weighting Paradigm.} This paradigm originates from the BM25~\cite{BM25} algorithm for exact lexicon matching, which utilizes an inverted index to reduce retrieval latency by only considering samples with overlapping lexicons during the retrieval process. This method has recently gained popularity in NLP document retrieval~\cite{splade2,splade,splade3,lexmae,unified}. However, while the text data is naturally discrete and can be projected into a vocabulary space, images are continuous and pose a challenge for sparse lexicon representation. Our proposed \textbf{LexLIP} is the first work to address this challenge in ITR.

\textbf{Bottlenecked Pre-training in Retrieval.} This method is widely studied in the document retrieval~\cite{condensor,cocondensor,retromae,lexmae,simlm}. The masked language modeling objective is conditioned on dense representations. Despite its proven success in NLP, this method has not yet been explored in ITR. In this work, we aim to fill this gap by proposing a novel pre-training method, which leverages sparse lexicon representations as bottlenecks to enhance the performance of our ITR model.

%% file: figure_mae.tex
\begin{figure*}
    \centering
    \includegraphics[height=10cm]{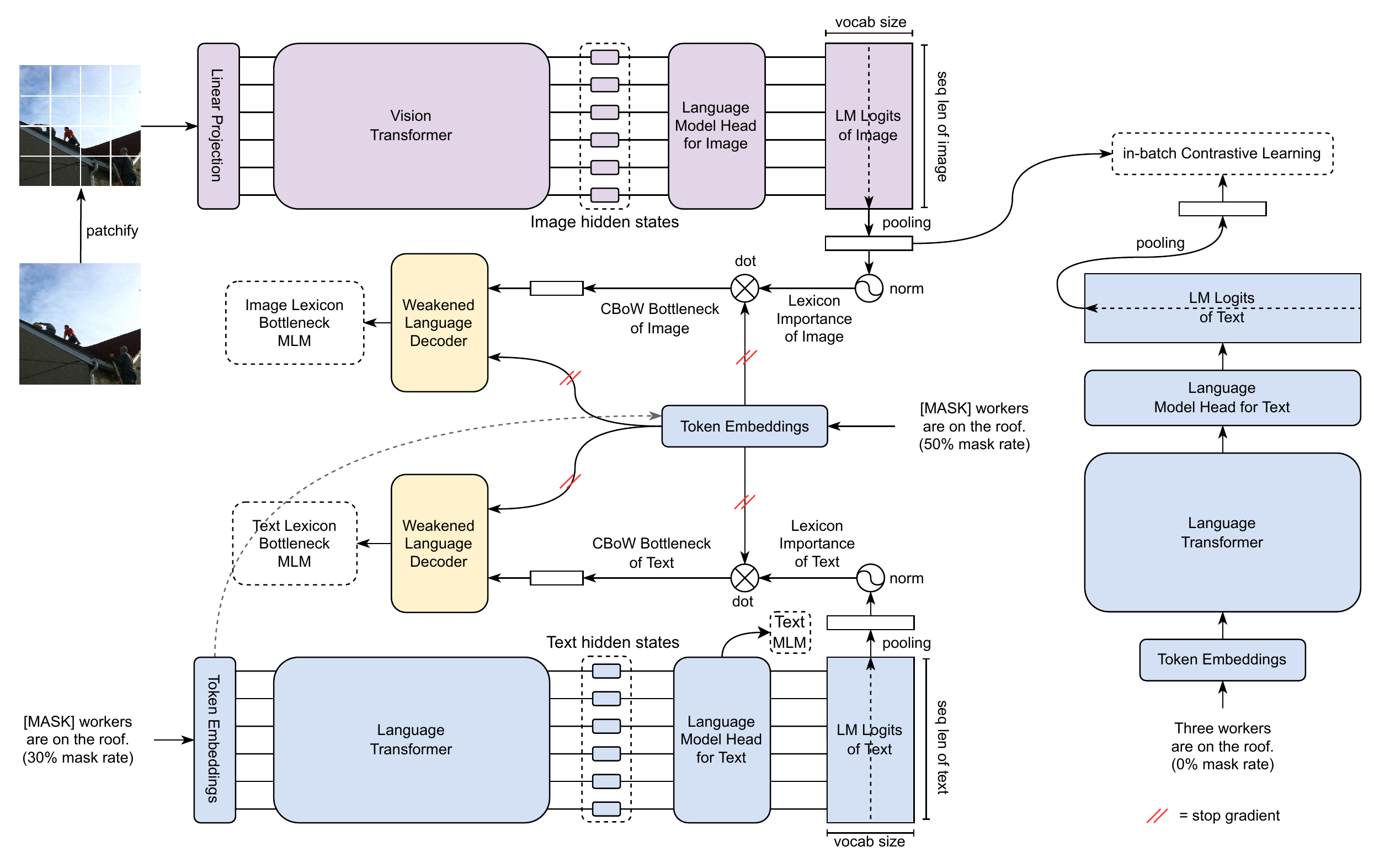}
    \caption{An overview of the Lexicon-Bottlenecked Pre-training phase, including self-supervised masked language modeling, image/text lexicon-bottlenecked masked language modeling, and in-batch lexicon-contrastive learning.}
    \label{fig:lexmae}
\end{figure*}

%% file: figure_lexmoco.tex
\begin{figure}
    \centering
    \includegraphics[height=4.8cm]{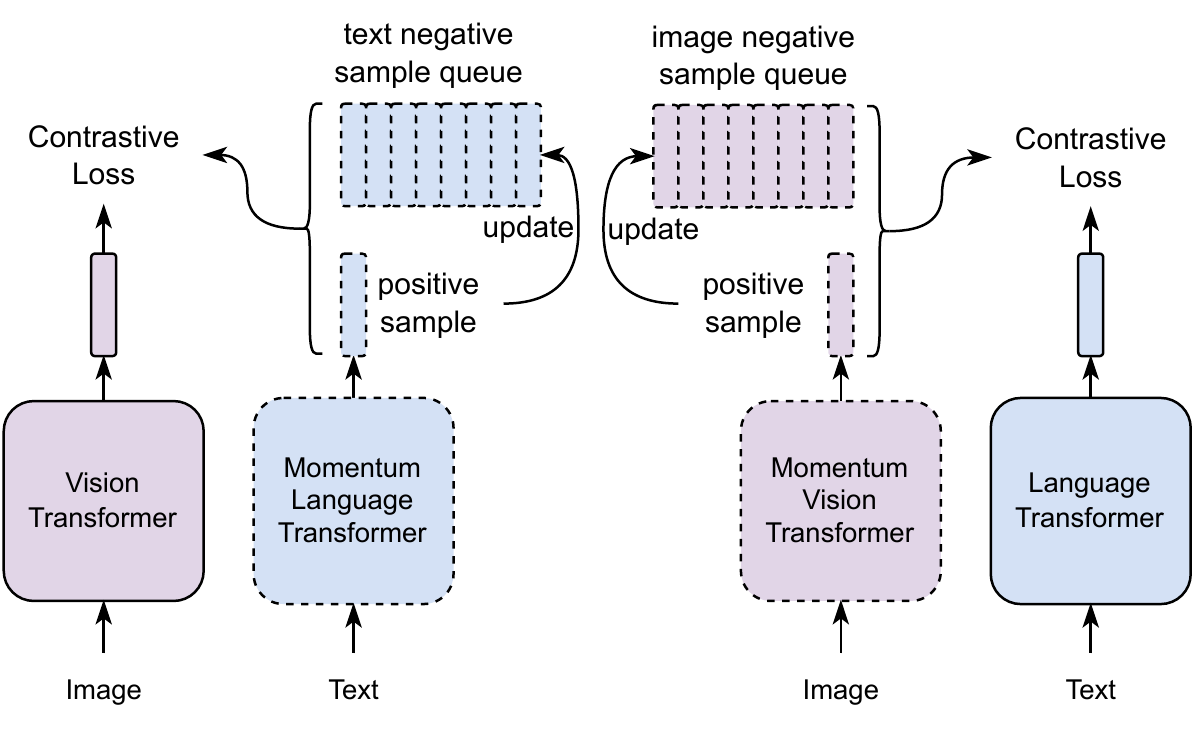}
    \caption{An overview of the Momentum Lexicon-Contrastive Pre-training phase.}
    \label{fig:lexmoco}
\end{figure}

%% file: 3.Methodology.tex
\section{LexLIP}

\textbf{Overview.} As depicted in Figures~\ref{fig:lexmae} and~\ref{fig:lexmoco}, our \textbf{LexLIP} pre-training framework comprises three crucial components, specifically: (i) Dual-Stream Encoders, (ii) Lexicon-Bottlenecked Modules, and (iii) Weakened Masking-Style Text Decoders. Furthermore, this framework consists of two pre-training phases, namely: (i) Lexicon-Bottlenecked Pre-training, and (ii) Momentum Lexicon-Contrastive Pre-training. In the following section, we will delve into the pre-training and inference processes of our framework.

\subsection{Dual-Stream Encoders and Sparse Lexicon Representations}

Adhering to the recent trends in ITR~\cite{align,cots,clip,sun-etal-2021-lightningdot}, our \textbf{LexLIP} framework employs dual-stream encoders to embed both images and texts into distinct representations. The backbone of the visual encoder is the Vision Transformer~\cite{vit}, while the Language Transformer~\cite{bert} serves as the backbone of the textual encoder. The input image is first transformed into a series of flattened 2D patches, which are subsequently processed by the visual encoder to generate the corresponding hidden states. Formally, given all patches of an image $x=[x_1,\dots,x_m]$, the visual encoder transform them into fixed-length vectors:
\begin{equation}
    H^v = \text{Trans}^{v}\left(\text{[CLS}^v;x\text{]}\right)\in \mathbb{R}^{(m+1)\times d},
\end{equation}
where $\text{Trans}^{v}$ is the visual encoder and $d$ is the model size. In the dense retrieval paradigm, it is a common practice to utilize the first hidden state of $H^v$ as the dense representations of images~\cite{cots,conlip,clip}. Differently, our visual encoder is followed by a language model head which projects the hidden states into the sparse vocabulary space:
\begin{equation}\label{eq:imagelogits}
    \bm{S}^{\text{(enc)}^{v}}_x = \text{LM-Head}^v(H^v)\in \mathbb{R}^{(m+1)\times \mathbb{|V|}},
\end{equation}
where $\mathbb{|V|}$ is the vocabulary size. We denote $\bm{S}^{\text{(enc)}^{v}}_x$ as the LM logits of images from visual encoder. Then, we follow the SPLADE model in document retrieval~\cite{splade} to represent an image in the high-dimensional vocabulary space by
\begin{equation}\label{eq:imagerep}
    p^v=\log(1+\text{MaxPool}(\max(\bm{S}^{\text{(enc)}^{v}}_x, 0)))\in \mathbb{R}^{\mathbb{|V|}},
\end{equation}
where $\max(\cdot, 0)$ ensures all values greater than or equal to zero for the sparse requirements, $\text{MaxPool}(\cdot)$ denotes max pooling along with the sequence axis, and the saturation function $\log(1 + \text{MaxPool}(\cdot))$ prevents some terms from dominating. $p^v$ stands for the lexicon-weighting sparse representation of an image.

Similarly, the language encoder generates the lexicon-weighting sparse representation of the input text $y=[y_1,...,y_n]$ by
\begin{equation}
    H^l = \text{Trans}^{l}\left(\text{[CLS}^l;y\text{]}\right)\in \mathbb{R}^{(n+1)\times d},
\end{equation}
\begin{equation}\label{eq:textlogits}
    \bm{S}^{\text{(enc)}^{l}}_y = \text{LM-Head}^l(H^l)\in \mathbb{R}^{(m+1)\times \mathbb{|V|}},
\end{equation}
\begin{equation}\label{eq:textrep}
    p^l=\log(1+\text{MaxPool}(\max(\bm{S}^{\text{(enc)}^{l}}_y, 0)))\in \mathbb{R}^{\mathbb{|V|}},
\end{equation}
where $\text{Trans}^{l}$ is the language encoder, $\bm{S}^{\text{(enc)}^{l}}_y$ is the LM logits of texts from language encoder, $\text{LM-Head}^l$ is the language model head for texts, and $p^l$ is the lexicon-weighting sparse representation of a text.

\subsection{Phase 1: Lexicon-Bottlenecked Pre-training}

As shown in Figure~\ref{fig:lexmae}, this pre-training phase consists of four different objectives, including self-supervised masked language modeling, two lexicon-bottlenecked masked language modelings and in-batch lexicon-contrastive learning.

\textbf{Self-Supervised Masked Language Modeling (SelfMLM).} Consistent with the standard practice of pre-training the language encoder in an unsupervised manner, the masked language modeling (MLM) objective is utilized for pre-training our language encoder, $\text{Trans}^{l}$. Formally, the tokens in the input text $y$ are masked to obtain $\bar{y}$, with $\alpha\%$ tokens being replaced by a special token [MASK] or a random token in the vocabulary set, $\mathbb{V}$, and the remaining being kept unchanged. The masked $\bar{y}$ is then processed by the language encoder to generate the language model (LM) logits, $\bm{S}^{\text{(enc)}^l}_{\bar{y}}$, and reconstruct the masked tokens through the following objective function:
\begin{equation}
    \mathcal{L}_{\text{self}}=-\sum_{\mathbb{D}}\sum_{j\in \mathbb{M}^{\text{(enc)}}}\log P(\textnormal{w}^j=y_j|\bar{y}),
\end{equation}
where $P(\textnormal{w}^j)$ is calculated as $softmax\left(\bm{S}^{\text{(enc)}^l}_{\bar{y}}[j,:]\right)$, $\mathbb{D}$ represents the set of all samples, $\mathbb{M}^{\text{(enc)}}$ denotes the set of masked indices in $\bar{y}$, $\textnormal{w}^j$ represents the discrete variable over $\mathbb{V}$ at the j-th position of $y$, and $y_j$ refers to its original token.

\textbf{Lexicon-Bottlenecked Masked Language Modelings (LexMLM).} Regarding to the token-level logits from Eq.~\ref{eq:imagelogits} and~\ref{eq:textlogits} defined in the lexicon vocabulary space, we propose to calculate the lexicon-importance distributions of images and masked texts by
\begin{equation}\label{eq:normalize_V}
a^v=\text{Normalize}\left(\text{MaxPool}(\bm{S}^{\text{(enc)}^v}_x)\right)\in[0,1]^{|\mathbb{V}|},
\end{equation}
\begin{equation}\label{eq:normalize_L}
a^l=\text{Normalize}\left(\text{MaxPool}(\bm{S}^{\text{(enc)}^l}_{\bar{y}})\right)\in[0,1]^{|\mathbb{V}|},
\end{equation}
where $\text{Normalize}(\cdot)=softmax(\cdot)$ denotes the normalization function (let $\sum a_i = 1$). $a^{(\cdot)}$ denotes lexicon-importance distribution over $\mathbb{V}$ to indicate the relative importance of the different lexicons in the vocabulary.

To obtain the lexicon-importance distributions, we are inspired by the bottleneck-enhanced dense representation learning strategy from recent works in document retrieval~\cite{condensor,cocondensor,retromae,lexmae}. Our framework utilizes these distributions as a bridge to guide the reconstruction of masked lexicons, leading the vision and language encoders to focus on the most critical tokens/words in the data. However, directly utilizing the high-dimensional distribution vectors $a^{(\cdot)}\in[0,1]^{|\mathbb{V}|}$ as the bottlenecks faces challenges. First, the distribution over the whole $\mathbb{V}$ has the ability to contain most semantics of data~\cite{DBLP:conf/iclr/YangDSC18}, making the bottleneck less effective. Second, the high-dimensional vector is difficult to feed into a decoder for representation learning and text reconstruction.

Therefore, we present a novel approach in which we generate continuous bag-of-words (CBoW) representations as the bottlenecks, informed by the lexicon-importance distributions obtained from Eq.~\ref{eq:normalize_V} and~\ref{eq:normalize_L}. That is
\begin{equation}
    b^{(\cdot)}=a^{(\cdot)}sg\left(W^{\text{(te)}}\right)\in \mathbb{R}^d,
\end{equation}
where $W^{\text{(te)}}\in R^{|\mathbb{V}|\times d}$ is the token embeddings matrix of the language encoder and $sg(\cdot)$ refers to stop gradient. Thereby, $b^{(\cdot)}$ stands for \textbf{CBoW bottleneck representations}.

To guide the learning of the bottleneck representations $b^{(\cdot)}$, which in turn leads to the learning of the lexicon-importance distributions $a^{(\cdot)}$, we use two decoders, one for vision and one for language, to reconstruct the masked text $\bar{y}$ from $b^{(\cdot)}$. This approach follows recent advancements in bottleneck-enhanced neural structures~\cite{condensor,cocondensor,lexmae}. The two decoders, which we refer to as \textbf{weakened masking-style decoders}, are designed to place a heavy reliance on the bottleneck representations by employing two strategies: (i) an aggressive masking strategy, and (ii) using only two shallow transformer layers.

In particular, given the masked text $\bar{y}$, we adopt an aggressive masking strategy to produce the masked text $\hat{y}$ with a larger masking rate. This prompts the encoders to compress the rich contextual information into the bottleneck representation, $b^{(\cdot)}$. Subsequently, the bottleneck representation prefixes $\hat{y}$ by replacing the [CLS] special token. Therefore, our weakened masking-style decoding can be formulated as
\begin{equation}
    \bm{S}^{\text{(dec)}^v}_{\hat{y}} = \text{Decoder}^v\left([b^v;\hat{y}]\right)\in \mathbb{R}^{(n+1)\times \mathbb{|V|}},
\end{equation}
\begin{equation}
    \bm{S}^{\text{(dec)}^l}_{\hat{y}} = \text{Decoder}^l\left([b^l;\hat{y}]\right)\in \mathbb{R}^{(n+1)\times \mathbb{|V|}}.
\end{equation}
Similar to the Self-MLM, the loss functions are:
\begin{equation}
    \mathcal{L}_{(i2t)}=-\sum_{\mathbb{D}}\sum_{j\in \mathbb{M}^{\text{(dec)}}}\log P^v(\textnormal{w}^j=y_j|\hat{y}),
\end{equation}
\begin{equation}
    \mathcal{L}_{t2i}=-\sum_{\mathbb{D}}\sum_{j\in \mathbb{M}^{\text{(dec)}}}\log P^l(\textnormal{w}^j=y_j|\hat{y}),
\end{equation}
where $P^{(\cdot)}(\textnormal{w}^j)=softmax\left(\bm{S}^{\text{(dec)}^{(\cdot)}}_{\hat{y}}[j,:]\right)$, and $\mathbb{M}^{\text{(dec)}}$ denotes the set of masked tokens in $\hat{y}$.

\textbf{In-Batch Lexicon-Contrastive Learning (BaCo).} Given the lexicon sparse representations from Eqs.~\ref{eq:imagerep} and~\ref{eq:textrep}, we perform in-batch contrastive learning in this phase to align images and texts in the vocabulary space. The models learn by contrasting the lexicon-
weighting sparse representations of different samples within a single batch of data. The loss functions are:
\begin{equation}\label{eq:i2tbaco}
    \mathcal{L}_{baco}^{i2t}=-\sum_{\mathbb{D}}\log\frac{\exp\left(p^v(p^l)^T\right)/{\tau}}{\sum_{j\in \mathcal{B}}\exp\left(p^v(p^l_j)^T\right)/{\tau}}+\lambda \mathcal{F}(p^v),
\end{equation}
\begin{equation}\label{eq:t2ibaco}
    \mathcal{L}_{baco}^{t2i}=-\sum_{\mathbb{D}}\log\frac{\exp\left(p^v(p^l)^T\right)/{\tau}}{\sum_{j\in \mathcal{B}}\exp\left(p^v_{j}(p^l)^T\right)/{\tau}}+\lambda \mathcal{F}(p^l),
\end{equation}
where $\mathcal{B}$ denotes all the data in a batch, $\tau$ is the temperature hyperparameter, $\mathcal{F}(\cdot)$ is the FLOPS function introduced in SPLADE~\cite{splade} for representation sparsity, and $\lambda$ is the regularization hyperparameter. The overall loss function is:
\begin{equation}
\mathcal{L}_{baco}=\left(\mathcal{L}_{baco}^{i2t} + \mathcal{L}_{baco}^{t2i}\right) / 2.
\end{equation}

\textbf{Phase 1 Learning.} The final loss function of the Lexicon-Bottlenecked Pre-training is a direct addition of all losses:
\begin{equation}
    \mathcal{L}_{p1}=\mathcal{L}_{self} + \mathcal{L}_{i2t} + \mathcal{L}_{t2i} + \mathcal{L}_{baco}.
\end{equation}

\input{compare_table.tex}

\subsection{Phase 2: Momentum Lexicon-Contrastive Pre-training.}

After learning the sparse lexicon representations in the first phase, we further align the representations of images and texts in the vocabulary space. It has been shown that the large-scale negative samples is crucial for achieving good performance in ITR~\cite{clip}. However, the negative sample size is limited by the mini-batch size in traditional in-batch contrastive learning, which can be constrained by the GPU's memory. To address this issue, we adopt the momentum contrastive learning in MoCo~\cite{moco} to cache negative samples with two different queues, $Q^v$ and $Q^l$, for images and texts, respectively. This approach decouples the negative sample size from the mini-batch size, making the learning process more computationally feasible.

In accordance with prior works~\cite{cots,conlip}, two momentum encoders, $\theta_{m}^{v}$ and $\theta_{m}^{l}$, are employed to update the samples in the queues. These encoders share the same structures and initial parameters as the original encoders, but they have truncated gradients and are updated utilizing the exponential moving average (EMA) mechanism:
\begin{equation}
    \theta_{m}^{v} = m \theta_{m}^{v} + (1 - m)\theta_{o}^{v},
\end{equation}
\begin{equation}
    \theta_{m}^{l} = m \theta_{m}^{l} + (1 - m)\theta_{o}^{l},
\end{equation}
where $\theta_{o}$ is the parameters of the original encoders and $m$ is the EMA decay weight. The momentum lexicon sparse representations of images and texts are denoted as $\hat{p}^{v}$ and $\hat{p}^{l}$. These momentum encoders are dropped after pre-training. The momentum contrastive loss functions are:
\begin{equation}\label{eq:i2tmoco}
    \begin{aligned}
        \mathcal{L}_{moco}^{i2t}=&-\sum_{\mathbb{D}}\log\frac{\exp\left(p^v(\hat{p}^l)^T\right)/{\tau}} {\sum_{q_j^l\in Q^l\cup \{\hat{p}^l\}}\exp\left(p^{v}(q_j^l)^T\right)/{\tau}},\\
        &+\lambda \mathcal{F}(p^v),
    \end{aligned}
\end{equation}
\begin{equation}\label{eq:t2imoco}
    \begin{aligned}
        \mathcal{L}_{moco}^{t2i}=&-\sum_{\mathbb{D}}\log\frac{\exp\left(p^l(\hat{p}^v)^T\right)/{\tau}} {\sum_{q_j^v\in Q^v\cup \{\hat{p}^v\}}\exp\left(p^{l}(q_j^v)^T\right)/{\tau}},\\
        &+\lambda \mathcal{F}(p^l).
    \end{aligned}
\end{equation}
The overall momentum contrastive loss is:
\begin{equation}
    \mathcal{L}_{moco} = (\mathcal{L}_{moco}^{i2t} + \mathcal{L}_{moco}^{t2i}) / 2.
\end{equation}

\subsection{Exact Lexicon Search for Large-Scale Retrieval}

In the inference phase of large-scale retrieval, there are distinct differences between the dense-vector and lexicon-weighting retrieval methods. As in Eq.~\ref{eq:i2tbaco},~\ref{eq:t2ibaco},~\ref{eq:i2tmoco}, and~\ref{eq:t2imoco}, we use the dot-product between the real-valued sparse lexicon-weighted representations to measure the similarity, where ``real-valued" is a requirement for gradient back-propagation and end-to-end learning. However, it is infeasible for practical purposes, especially for open-source term-based retrieval systems, such as Anserini~\cite{anserini}. To overcome this challenge, we employ ``quantization" and ``term-based systems" to convert the high-dimensional sparse vectors back to the corresponding lexicons and their virtual frequencies (weights). The lexicons are derived from the non-zero elements in the high-dimensional sparse vector, and the weights are obtained through straightforward quantization (i.e., $\lfloor100 \times p^{(\cdot)}\rfloor$). Given a query and a candidate sample, the exact lexicon matching score is defined as:
\begin{equation}
    score = \sum_{l\in L^{q}\cap L^{s}} W_{q}(l)\times W_{s}(l),
\end{equation}
where $L^{q}$ and $L^{s}$ denote the lexicon lists of the query and candidate sample, and $W_{(\cdot)}(l)$ is the weight of the lexicon.

\input{compare_qps_table.tex}

Overall, our large-scale retrieval framework, \textbf{LexLIP}, is comprised of the following steps: i) converting candidate samples into high-dimensional sparse vectors and subsequently into lexicons and frequencies, ii) constructing a term-based inverted index using Anserini~\cite{anserini} for the entire collection, iii) generating the lexicons and frequencies for a test query similarly, and iv) querying the index to retrieve the top candidates.

%% file: compare_table.tex
\begin{table*}[h!]
    \footnotesize
    \centering
    \begin{tabular}{lccccccccccccc}
        \toprule
        \multirow{3}{*}{\textbf{Model}} & \multirow{3}{*}{*\textbf{\#I-T}} & \multicolumn{6}{c}{\textbf{Flickr30k Test (1K Images)}} & \multicolumn{6}{c}{\textbf{MSCOCO Test (5K Images)}}\\
        & & \multicolumn{3}{c}{\textbf{T2I Retrieval}} & \multicolumn{3}{c}{\textbf{I2T Retrieval}} & \multicolumn{3}{c}{\textbf{T2I Retrieval}} & \multicolumn{3}{c}{\textbf{I2T Retrieval}}\\
        & & R@1 & R@5 & R@10 & R@1 & R@5 & R@10 & R@1 & R@5 & R@10 & R@1 & R@5 & R@10\\
        \hline\hline
        \multicolumn{10}{l}{\textit{Dense-vector Dual-Stream Retriever}}\\
        \hline
        \textbf{Frozen} (ICCV21~\cite{DBLP:conf/iccv/BainNVZ21}) & 5.5M & 61.0 & 87.5 & 92.7 & - & - & - & - & - & - & - & - & -\\
        \textbf{LightDOT} (NAACL21~\cite{sun-etal-2021-lightningdot}) & 9.5M & 69.9 & 91.1 & 95.2 & 83.9 & 97.2 & 98.6 & 45.8 & 74.6 & 83.8 & 60.1 & 85.1 & 91.8\\
        \textbf{COOKIE} (ICCV21~\cite{cookie}) & 5.9M & 68.3 & 91.1 & 95.2 & 84.7 & 96.9 & 98.3 & 46.6 & 75.2 & 84.1 & 61.7 & 86.7 & 92.3\\
        \textbf{ViSTA} (CVPR22~\cite{vista}) & 9.5M & 68.9 & 91.1 & 95.1 & 84.8 & 97.4 & 99.0 &  47.8 & 75.8 & 84.5 & 63.9 & 87.8 & 93.6\\
        $\dagger$\textbf{Dense} (ours) & 4.3M & 74.0 & 92.8& 95.6 & 88.0 & 98.1 & 99.5 & 49.5 & 76.7 & 85.4 & 65.5 & 88.6 & 94.3\\
        $\ddagger$\textbf{COTS} (CVPR22~\cite{cots}) & 5.3M & 75.2 & 93.6 & 96.5 & 88.2 & 98.5 & 99.7 & 50.5 & 77.6 & 86.1 & 66.9 & 88.8 & 94.0\\
        $\ddagger$\textbf{COTS} (CVPR22~\cite{cots}) & 15.3M & 76.5 & 93.9 & 96.6 & 90.6 & 98.7 & 99.7 & 52.4 & 79.0 & 86.9 & 69.0 & 90.4 & 94.9\\
        \hline\hline
        \multicolumn{10}{l}{\textit{Sparse-vector Dual-Stream Retriever}}\\
        \hline
        \textbf{LexLIP} (ours) & 4.3M & 76.7 & 93.7 & 96.8 & 89.6 & 98.7 & 99.6 & 51.9 & 78.3 & 86.3 & 67.9 & 89.7 & 94.8\\
        \textbf{LexLIP} (ours) & 14.3M & \textbf{78.4} & \textbf{94.6} & \textbf{97.1} & \textbf{91.4} & \textbf{99.2} & \textbf{99.7} & \textbf{53.2} & \textbf{79.1} & 86.7 & \textbf{70.2} & \textbf{90.7} & \textbf{95.2}\\
        \bottomrule
    \end{tabular}
    \begin{tablenotes}
        \item{*} \small{\text{\#I-T corresponds to the number of image-text pairs during pre-training.}}
        \item{$\dagger$} \small{\text{Replace the sparse representations with the dense one in our framework.}}
        \item{$\ddagger$} \small{\text{The state-of-the-art dense-vector dual-stream retriever, under fair comparison.}}
    \end{tablenotes}
    \caption{Evaluation our \textbf{LexLIP} in the small-scale retrieval scenario after fine-tuning.}
    \label{tab:compare}
\end{table*}

%% file: compare_qps_table.tex
\begin{table*}
    \footnotesize
    \centering
    \begin{tabular}{lccccccccc}
        \toprule
        \multirow{3}{*}{\textbf{Model}} & \multirow{3}{*}{\textbf{Index Size}$\downarrow$} & \multirow{3}{*}{\textbf{Repr Byte}$\downarrow$} &\multirow{3}{*}{\textbf{QPS}$\uparrow$} & \multicolumn{6}{c}{\textbf{Large-Scale Flickr30k Test}}\\
        & & & & \multicolumn{3}{c}{\textbf{T2I Retrieval}} & \multicolumn{3}{c}{\textbf{I2T Retrieval}}\\
        & & & & R@1 & R@5 & R@10 & R@1 & R@5 & R@10\\
        \hline
        \hline
        \multicolumn{6}{l}{\textit{Single-Modal Sparse Text Retriever}}\\
        \hline
        ~\textbf{BM25}~\cite{BM25} & 195M & Avg 195 & 780.18 & 16.8 & 27.3 & 31.9 & 34.0 & 50.7 & 58.2\\
        \hline
        \hline
        \multicolumn{6}{l}{\textit{Cross-Modal Dense Dual-Stream Retriever}}\\
        \hline
        ~\textbf{CLIP}~\cite{clip} & 2.0G & Avg 1998 & 3.60 & 14.0 & 28.7 & 36.6 & 35.4 & 59.9 & 70.0\\
        \hline
        \hline
        \multicolumn{6}{l}{\textit{Our Cross-Modal Sparse Dual-Stream Retriever}}\\
        \hline
        ~\textbf{LexLIP} & 152M & Avg 152 & 19.70 & 48.6 & 66.8 & 71.8 & 64.1 & 86.5 & 91.4\\
        ~~-top64 sparsify & 114M & Upto192 & 63.43 & 47.6 & 65.7 & 70.6 & 64.4 & 86.7 & 91.1\\
        ~~-top32 sparsify & 71M & Upto 96 & 249.06 & 42.0 & 59.3 & 64.7 & 59.4 & 83.4 & 87.7\\
        ~~-top16 sparsify & 47M & Upto 48 & 610.08 & 28.3 & 44.2 & 50.6 & 47.4 & 70.1 & 75.9\\
        ~~-top12 sparsify & 41M & Upto 36 & 796.65 & 22.3 & 35.9 & 41.4 & 40.9 & 60.4 & 70.2\\
        ~~-top8 sparsify & 34M & Upto 24 & 985.40 & 14.6 & 24.7 & 29.5 & 28.6 & 47.2 & 55.5\\ 
        \hline
    \end{tabular}
    \caption{Evaluating our \textbf{LexLIP} in the large-scale retrieval scenario. ``Index Size'' corresponds to the storage requirement to embedded all candidate images. ``Repr Byte'' denotes the storage requirement for an embedded image. Each activated (non-zero) term in lexicon-weighed sparse vector needs 3 bytes (2 bytes for indexing and 1 byte for its weight). ``QPS'' corresponds to query-per-second, which denotes the retrieval latency, the higher the faster.}
    \label{tab:large}
\end{table*}

%% file: 4.Experiment.tex
\section{Small-Scale Retrieval Experiment}

\subsection{Setup}

\textbf{Pre-Training and Evaluation.} We use two different image-text datasets to pre-train our \textbf{LexLIP}: (1) CC4.3M contains 4.3M image-text pairs from Conceptual Captions 3.3M~\cite{cc3m}  (about 2.8M urls are valid), SBU~\cite{sbu}, MSCOCO~\cite{coco} training set
and Flickr30K~\cite{f30k} training set. (2) CC14.3M consists of CC4.3M and
Conceptual Captions 12M~\cite{cc12m} (about 10M urls are valid), which
contains 14.3M image-text pairs. For the downstream tasks, models are evaluated on MSCOCO and Flickr30k with fine-tuning. Each image in these datasets is accompanied by 5 different captions. We follow the Karpathy split~\cite{DBLP:conf/cvpr/KarpathyL15} to divide the datasets into train/val/test sets, with 113.2k/5k/5k (MSCOCO) and 29.8k/1k/1k (Flickr30k) images. For evaluation, we use the standard R@k (k=1,5,10) to calculate the retrieval scores of our models on the test sets.

\textbf{Implementation Details.} For computational efficiency, we follow~\cite{cots} to initialize the dual-stream encoders with the pre-trained vision~\cite{beit} and language transformer~\cite{bert}, whereas the other parts are randomly initialized. Both of them are the base-size, 12-layer transformer encoder with 768 hidden size and 12 attention heads. The pre-trained input image resolution is $224\times 224$. The fine-tuning resolution is $384\times 384$. Models are pre-trained with 20 epochs in the first phase, 10 epochs in the second phase, and fine-tuned with 10 epochs. The AdamW optimization algorithm~\cite{adamw} with a learning rate of 5e-5, linear learning rate decay and 10\% warm-up steps, and mixed-precision training are employed. The masking rate for the language encoder is set to 30\% and 50\% for the decoder, with an EMA weight of 0.99 and a temperature $\tau$ of 0.05. The regularization term $\lambda$ is set to 0.002. Further details can be found in the supplementary materials.

\subsection{Results}

As shown in Table~\ref{tab:compare}, under a fair comparison (excluding the models pre-trained with billions of image-text pairs), our \textbf{LexLIP} achieves the SOTA performance over all previous works for most evaluation metrics. Specifically, in comparison to the previous SOTA COTS~\cite{cots}, \textbf{LexLIP} obtains higher results by 1.5\% (76.7\% vs. 75.2\%) for T2I R@1 and 1.4\% (89.6\% vs. 88.2\%) for I2T R@1 on Flickr30k, while utilizing less pre-training data (4.3M vs. 5.3M). Furthermore, with a larger pre-training dataset, \textbf{LexLIP} further enhances performance by 1.9\% (78.4\% vs. 76.5\%) for T2I R@1 and 0.8\% (91.4\% vs. 90.6\%) for I2T R@1 on Flickr30k, while still utilizing less data (14.3M vs. 15.3M).

\begin{figure*}
    \centering
    \includegraphics[height=4.5cm]{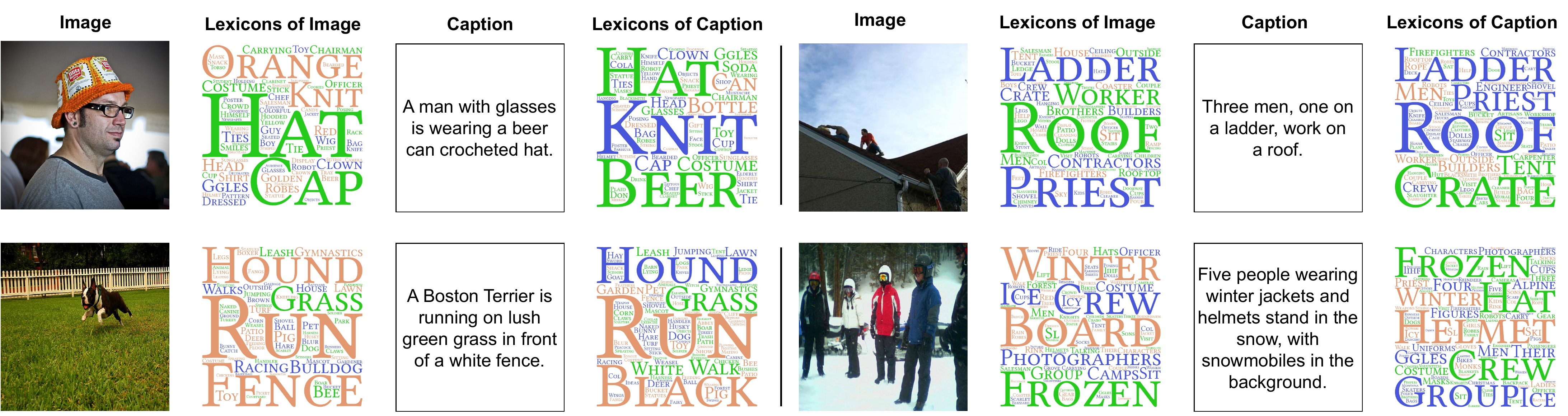}
    \caption{Visualizing the lexicons of the images and their corresponding captions. If a lexicon has a larger weight, it will be larger in the lexicon cloud.}
    \label{fig:cloud}
\end{figure*}

\input{compare_ablation.tex}

\section{Large-Scale Retrieval Experiment}

\subsection{Setup}

\textbf{Baselines.} We include the SOTA dense retriever, CLIP-B/16~\cite{clip} as our baseline, which is pre-trained with around 400M image-text pairs. Additionally, we employ the single-modal sparse text retriever, BM25~\cite{BM25}, as another baseline. We utilize captions to represent images, enabling BM25 to perform image-text retrieval as caption-text retrieval.

\textbf{Large-Scale Benchmark.} To the best of our knowledge, this benchmark has not been previously established. We chose to expand the test set of Flickr30k~\cite{f30k} by incorporating 1M random image-text pairs from Conceptual Caption 12M~\cite{cc12m}. Each image in Flickr30k is associated with 5 captions, from which we randomly select one as the alternative image representation for BM25 retrieval. In text-to-image retrieval, models retrieve images from the 1k images of Flickr30k and an additional 1M images, given 4k captions as queries from Flickr30k. Conversely, for image-to-text retrieval, models must retrieve captions from the 4k captions of Flickr30k and an additional 1M captions, given 1k images as queries from Flickr30k.

\textbf{Pre-training Datasets.} To compare with CLIP in the zero-shot settings, we exclude the Flickr30k training set from the CC4.3M and result in a new dataset CC4.2M for our \textbf{LexLIP} pre-training.

\textbf{Retrieval Details.} For all models, we first embed them into the Index file and then retrieve the results on CPU. The dense retrieval is conduct with the efficient dense vector similarity search library, Faiss~\cite{faiss}. The sparse retrieval is conduct with the efficient sparse vector similarity search library, Anserini~\cite{anserini}. More details can be found in the supplemental material.

\subsection{Results}

Given the significance of large-scale retrieval systems with millions or billions of samples in serving as an upstream module for downstream tasks or being a fundamental component in commercial applications, it is important to evaluate these systems based on key metrics that assess their efficiency, such as retrieval latency (measured in terms of query-per-second or QPS), the size of the inverted index, and the representation size per sample. 

Table~\ref{tab:large} reveals the inefficiency of the dense retriever CLIP, with a retrieval speed of 3.60 QPS, which is not suitable for real-life applications. In contrast, our sparse retriever, \textbf{LexLIP}, demonstrates a substantial improvement, with a more than 5 times faster retrieval speed (19.70 vs. 3.60) and more than 13 times less storage memory (152M vs. 2.0G). Additionally, our \textbf{LexLIP} also exhibits superior performance on the benchmark.

To further analyze our \textbf{LexLIP} efficacy-efficiency trade-off, we adopt a simple yet effective sparsification method was adopted by retaining only the top-K weighted terms in the representation of the sample during the inference phase and constructing the inverted index using the sparsified samples. As shown in Table~\ref{tab:large}, our \textbf{LexLIP} achieves the best efficacy-efficiency trade-off among all baselines. With top-12 sparsity, \textbf{LexLIP} has the 4.8 times smaller index size, faster QPS, and better retrieval performance than the sparse text retriever BM25. Compared to the dense retriever, CLIP, the index size is 48.8 times smaller and the QPS is 221.3 times faster of our \textbf{LexLIP}.

\section{Further Analysis}

\subsection{Ablation Study}

In Table~\ref{tab:ablation}, we present the impact of different pre-training objectives and phases of our \textbf{LexLIP}. A pre-training dataset was constructed by randomly sampling 1.4M image-text pairs from Conceptual Captions 3.3M~\cite{cc3m} and including all pairs from the Flickr30k~\cite{f30k} training set. The retrieval performance was evaluated on the Flickr30k test set. The results indicate that all pre-training objectives and phases contribute positively to the retrieval performance. The greatest effects were observed for the in-batch contrastive learning in Phase 1 and the momentum contrastive learning in Phase 2, which may be attributed to the alignment of these objectives with the retrieval target. The MLM-based objectives were also found to be beneficial for learning the lexicon-importance distributions.

\subsection{Lexicon-Weighting Examples}

As shown in Figure~\ref{fig:cloud}, we visualize the lexicons of 4 images and their captions in the Flickr30k test set. If the lexicon has a high weight, the size is large in the lexicon cloud. We can find that the major features of the images and texts are successfully captured by the lexicons. For example, ``hat'' is an important lexicon in the first image and caption.

%% file: compare_ablation.tex
\begin{table*}
    \footnotesize
    \centering
    \begin{tabular}{ccccccccccc}
        \toprule
        \multicolumn{4}{c}{\textbf{Phase 1}} & \textbf{Phase 2} & \multicolumn{3}{c}{\textbf{T2I Retrieval}} & \multicolumn{3}{c}{\textbf{I2T Retrieval}}\\
        $\mathcal{L}_{self}$ & $\mathcal{L}_{i2t}$ & $\mathcal{L}_{t2t}$ & $\mathcal{L}_{baco}$ & $\mathcal{L}_{moco}$ & R@1 & R@5 & R@10 & R@1 & R@5 & R@10\\
         \hline\hline
         \checkmark & \checkmark & \checkmark & \checkmark & \checkmark & 
         \textbf{67.5} & \textbf{89.7} & \textbf{94.1} & 81.3 & \textbf{96.4} & \textbf{98.9}\\
         \checkmark & \checkmark & \checkmark & \checkmark & &
         59.5 & 85.6 & 91.8 & 73.4 & 92.9 & 96.5\\
         \checkmark & \checkmark & \checkmark & & \checkmark & 
         61.0 & 86.7 & 92.1 & 72.7 & 92.6 & 96.6\\
         \checkmark & \checkmark & & \checkmark & \checkmark &
         66.5 & 89.4 & 93.8 & \textbf{81.7} & 95.0 & 97.9\\
         \checkmark & & \checkmark & \checkmark & \checkmark &
         66.9 & 89.6 & \textbf{94.1} & 80.8 & 95.5 & 98.7\\
         & \checkmark & \checkmark & \checkmark & \checkmark &
         67.0 & 89.3 & 93.5 & 78.7 & 95.2 & 97.8\\
         \bottomrule
    \end{tabular}
    \caption{\textbf{LexLIP} ablation experiments on different pre-training objectives.}
    \label{tab:ablation}
\end{table*}

%% file: 5.Conclusion.tex
\section{Conclusion}

In this study, we present the novel \textbf{Lexicon-Weighting Paradigm} for ITR. To address the challenge of reconciling continuous image data with discrete vocabulary space, we propose the \textbf{Lexicon-Bottlenecked Language-Image Pre-training (LexLIP)} framework. Our experimental results demonstrate that \textbf{LexLIP} outperforms the SOTA models on small-scale retrieval benchmarks when pre-trained with similarly sized data. Furthermore, in large-scale retrieval, \textbf{LexLIP} demonstrates substantial improvement in terms of both retrieval speed ($5.5\sim221.3$ times faster) and storage memory requirements ($13.2\sim48.8$ times less) compared to the dense retriever, CLIP.

%% file: egbib.bbl
\begin{thebibliography}{10}\itemsep=-1pt

\bibitem{DBLP:conf/iccv/BainNVZ21}
Max Bain, Arsha Nagrani, G{\"{u}}l Varol, and Andrew Zisserman.
\newblock Frozen in time: {A} joint video and image encoder for end-to-end
  retrieval.
\newblock In {\em 2021 {IEEE/CVF} International Conference on Computer Vision,
  {ICCV} 2021, Montreal, QC, Canada, October 10-17, 2021}, pages 1708--1718.
  {IEEE}, 2021.

\bibitem{beit}
Hangbo Bao, Li Dong, Songhao Piao, and Furu Wei.
\newblock Beit: {BERT} pre-training of image transformers.
\newblock In {\em The Tenth International Conference on Learning
  Representations, {ICLR} 2022, Virtual Event, April 25-29, 2022}.
  OpenReview.net, 2022.

\bibitem{CNN}
Yoshua Bengio, Yann LeCun, and Donnie Henderson.
\newblock Globally trained handwritten word recognizer using spatial
  representation, convolutional neural networks, and hidden markov models.
\newblock In Jack~D. Cowan, Gerald Tesauro, and Joshua Alspector, editors, {\em
  Advances in Neural Information Processing Systems 6, [7th {NIPS} Conference,
  Denver, Colorado, USA, 1993]}, pages 937--944. Morgan Kaufmann, 1993.

\bibitem{cc12m}
Soravit Changpinyo, Piyush Sharma, Nan Ding, and Radu Soricut.
\newblock Conceptual 12m: Pushing web-scale image-text pre-training to
  recognize long-tail visual concepts.
\newblock In {\em {IEEE} Conference on Computer Vision and Pattern Recognition,
  {CVPR} 2021, virtual, June 19-25, 2021}, pages 3558--3568. Computer Vision
  Foundation / {IEEE}, 2021.

\bibitem{vista}
Mengjun Cheng, Yipeng Sun, Longchao Wang, Xiongwei Zhu, Kun Yao, Jie Chen,
  Guoli Song, Junyu Han, Jingtuo Liu, Errui Ding, and Jingdong Wang.
\newblock Vista: Vision and scene text aggregation for cross-modal retrieval.
\newblock In {\em {IEEE/CVF} Conference on Computer Vision and Pattern
  Recognition, {CVPR} 2022, New Orleans, LA, USA, June 18-24, 2022}, pages
  5174--5183. {IEEE}, 2022.

\bibitem{bert}
Jacob Devlin, Ming{-}Wei Chang, Kenton Lee, and Kristina Toutanova.
\newblock {BERT:} pre-training of deep bidirectional transformers for language
  understanding.
\newblock In Jill Burstein, Christy Doran, and Thamar Solorio, editors, {\em
  Proceedings of the 2019 Conference of the North American Chapter of the
  Association for Computational Linguistics: Human Language Technologies,
  {NAACL-HLT} 2019, Minneapolis, MN, USA, June 2-7, 2019, Volume 1 (Long and
  Short Papers)}, pages 4171--4186. Association for Computational Linguistics,
  2019.

\bibitem{vit}
Alexey Dosovitskiy, Lucas Beyer, Alexander Kolesnikov, Dirk Weissenborn,
  Xiaohua Zhai, Thomas Unterthiner, Mostafa Dehghani, Matthias Minderer, Georg
  Heigold, Sylvain Gelly, Jakob Uszkoreit, and Neil Houlsby.
\newblock An image is worth 16x16 words: Transformers for image recognition at
  scale.
\newblock In {\em 9th International Conference on Learning Representations,
  {ICLR} 2021, Virtual Event, Austria, May 3-7, 2021}. OpenReview.net, 2021.

\bibitem{vse}
Fartash Faghri, David~J. Fleet, Jamie~Ryan Kiros, and Sanja Fidler.
\newblock {VSE++:} improving visual-semantic embeddings with hard negatives.
\newblock In {\em British Machine Vision Conference 2018, {BMVC} 2018,
  Newcastle, UK, September 3-6, 2018}, page~12. {BMVA} Press, 2018.

\bibitem{splade2}
Thibault Formal, Carlos Lassance, Benjamin Piwowarski, and St{\'{e}}phane
  Clinchant.
\newblock From distillation to hard negative sampling: Making sparse neural
  {IR} models more effective.
\newblock In Enrique Amig{\'{o}}, Pablo Castells, Julio Gonzalo, Ben
  Carterette, J.~Shane Culpepper, and Gabriella Kazai, editors, {\em {SIGIR}
  '22: The 45th International {ACM} {SIGIR} Conference on Research and
  Development in Information Retrieval, Madrid, Spain, July 11 - 15, 2022},
  pages 2353--2359. {ACM}, 2022.

\bibitem{splade}
Thibault Formal, Benjamin Piwowarski, and St{\'{e}}phane Clinchant.
\newblock {SPLADE:} sparse lexical and expansion model for first stage ranking.
\newblock In Fernando Diaz, Chirag Shah, Torsten Suel, Pablo Castells, Rosie
  Jones, and Tetsuya Sakai, editors, {\em {SIGIR} '21: The 44th International
  {ACM} {SIGIR} Conference on Research and Development in Information
  Retrieval, Virtual Event, Canada, July 11-15, 2021}, pages 2288--2292. {ACM},
  2021.

\bibitem{condensor}
Luyu Gao and Jamie Callan.
\newblock Condenser: a pre-training architecture for dense retrieval.
\newblock In Marie{-}Francine Moens, Xuanjing Huang, Lucia Specia, and
  Scott~Wen{-}tau Yih, editors, {\em Proceedings of the 2021 Conference on
  Empirical Methods in Natural Language Processing, {EMNLP} 2021, Virtual Event
  / Punta Cana, Dominican Republic, 7-11 November, 2021}, pages 981--993.
  Association for Computational Linguistics, 2021.

\bibitem{cocondensor}
Luyu Gao and Jamie Callan.
\newblock Unsupervised corpus aware language model pre-training for dense
  passage retrieval.
\newblock In Smaranda Muresan, Preslav Nakov, and Aline Villavicencio, editors,
  {\em Proceedings of the 60th Annual Meeting of the Association for
  Computational Linguistics (Volume 1: Long Papers), {ACL} 2022, Dublin,
  Ireland, May 22-27, 2022}, pages 2843--2853. Association for Computational
  Linguistics, 2022.

\bibitem{moco}
Kaiming He, Haoqi Fan, Yuxin Wu, Saining Xie, and Ross~B. Girshick.
\newblock Momentum contrast for unsupervised visual representation learning.
\newblock In {\em 2020 {IEEE/CVF} Conference on Computer Vision and Pattern
  Recognition, {CVPR} 2020, Seattle, WA, USA, June 13-19, 2020}, pages
  9726--9735. Computer Vision Foundation / {IEEE}, 2020.

\bibitem{DBLP:conf/wacv/HuVH22}
Brian Hu, Bhavan Vasu, and Anthony Hoogs.
\newblock {X-MIR:} explainable medical image retrieval.
\newblock In {\em {IEEE/CVF} Winter Conference on Applications of Computer
  Vision, {WACV} 2022, Waikoloa, HI, USA, January 3-8, 2022}, pages 1544--1554.
  {IEEE}, 2022.

\bibitem{align}
Chao Jia, Yinfei Yang, Ye Xia, Yi{-}Ting Chen, Zarana Parekh, Hieu Pham,
  Quoc~V. Le, Yun{-}Hsuan Sung, Zhen Li, and Tom Duerig.
\newblock Scaling up visual and vision-language representation learning with
  noisy text supervision.
\newblock In Marina Meila and Tong Zhang, editors, {\em Proceedings of the 38th
  International Conference on Machine Learning, {ICML} 2021, 18-24 July 2021,
  Virtual Event}, volume 139 of {\em Proceedings of Machine Learning Research},
  pages 4904--4916. {PMLR}, 2021.

\bibitem{faiss}
Jeff Johnson, Matthijs Douze, and Herv{\'{e}} J{\'{e}}gou.
\newblock Billion-scale similarity search with gpus.
\newblock {\em {IEEE} Trans. Big Data}, 7(3):535--547, 2021.

\bibitem{DBLP:conf/cvpr/KarpathyL15}
Andrej Karpathy and Li Fei{-}Fei.
\newblock Deep visual-semantic alignments for generating image descriptions.
\newblock In {\em {IEEE} Conference on Computer Vision and Pattern Recognition,
  {CVPR} 2015, Boston, MA, USA, June 7-12, 2015}, pages 3128--3137. {IEEE}
  Computer Society, 2015.

\bibitem{splade3}
Carlos Lassance and St{\'{e}}phane Clinchant.
\newblock An efficiency study for {SPLADE} models.
\newblock In Enrique Amig{\'{o}}, Pablo Castells, Julio Gonzalo, Ben
  Carterette, J.~Shane Culpepper, and Gabriella Kazai, editors, {\em {SIGIR}
  '22: The 45th International {ACM} {SIGIR} Conference on Research and
  Development in Information Retrieval, Madrid, Spain, July 11 - 15, 2022},
  pages 2220--2226. {ACM}, 2022.

\bibitem{DBLP:conf/kdd/LiLJLYZWM21}
Sen Li, Fuyu Lv, Taiwei Jin, Guli Lin, Keping Yang, Xiaoyi Zeng, Xiao{-}Ming
  Wu, and Qianli Ma.
\newblock Embedding-based product retrieval in taobao search.
\newblock In Feida Zhu, Beng~Chin Ooi, and Chunyan Miao, editors, {\em {KDD}
  '21: The 27th {ACM} {SIGKDD} Conference on Knowledge Discovery and Data
  Mining, Virtual Event, Singapore, August 14-18, 2021}, pages 3181--3189.
  {ACM}, 2021.

\bibitem{coco}
Tsung{-}Yi Lin, Michael Maire, Serge~J. Belongie, James Hays, Pietro Perona,
  Deva Ramanan, Piotr Doll{\'{a}}r, and C.~Lawrence Zitnick.
\newblock Microsoft {COCO:} common objects in context.
\newblock In David~J. Fleet, Tom{\'{a}}s Pajdla, Bernt Schiele, and Tinne
  Tuytelaars, editors, {\em Computer Vision - {ECCV} 2014 - 13th European
  Conference, Zurich, Switzerland, September 6-12, 2014, Proceedings, Part
  {V}}, volume 8693 of {\em Lecture Notes in Computer Science}, pages 740--755.
  Springer, 2014.

\bibitem{retromae}
Zheng Liu and Yingxia Shao.
\newblock Retromae: Pre-training retrieval-oriented transformers via masked
  auto-encoder.
\newblock {\em CoRR}, abs/2205.12035, 2022.

\bibitem{adamw}
Ilya Loshchilov and Frank Hutter.
\newblock Decoupled weight decay regularization.
\newblock In {\em 7th International Conference on Learning Representations,
  {ICLR} 2019, New Orleans, LA, USA, May 6-9, 2019}. OpenReview.net, 2019.

\bibitem{cots}
Haoyu Lu, Nanyi Fei, Yuqi Huo, Yizhao Gao, Zhiwu Lu, and Ji{-}Rong Wen.
\newblock {COTS:} collaborative two-stream vision-language pre-training model
  for cross-modal retrieval.
\newblock In {\em {IEEE/CVF} Conference on Computer Vision and Pattern
  Recognition, {CVPR} 2022, New Orleans, LA, USA, June 18-24, 2022}, pages
  15671--15680. {IEEE}, 2022.

\bibitem{conlip}
Ziyang Luo, Yadong Xi, Rongsheng Zhang, GongZheng Li, Zeng Zhao, and Jing Ma.
\newblock Conditioned masked language and image modeling for image-text dense
  retrieval.
\newblock In {\em Findings of the Association for Computational Linguistics:
  EMNLP 2022}, pages 130--140, Abu Dhabi, United Arab Emirates, Dec. 2022.
  Association for Computational Linguistics.

\bibitem{doc2query}
Rodrigo Nogueira, Jimmy Lin, and AI Epistemic.
\newblock From doc2query to doctttttquery.
\newblock {\em Online preprint}, 6, 2019.

\bibitem{sbu}
Vicente Ordonez, Girish Kulkarni, and Tamara~L. Berg.
\newblock Im2text: Describing images using 1 million captioned photographs.
\newblock In John Shawe{-}Taylor, Richard~S. Zemel, Peter~L. Bartlett, Fernando
  C.~N. Pereira, and Kilian~Q. Weinberger, editors, {\em Advances in Neural
  Information Processing Systems 24: 25th Annual Conference on Neural
  Information Processing Systems 2011. Proceedings of a meeting held 12-14
  December 2011, Granada, Spain}, pages 1143--1151, 2011.

\bibitem{f30k}
Bryan~A. Plummer, Liwei Wang, Chris~M. Cervantes, Juan~C. Caicedo, Julia
  Hockenmaier, and Svetlana Lazebnik.
\newblock Flickr30k entities: Collecting region-to-phrase correspondences for
  richer image-to-sentence models.
\newblock In {\em 2015 {IEEE} International Conference on Computer Vision,
  {ICCV} 2015, Santiago, Chile, December 7-13, 2015}, pages 2641--2649. {IEEE}
  Computer Society, 2015.

\bibitem{clip}
Alec Radford, Jong~Wook Kim, Chris Hallacy, Aditya Ramesh, Gabriel Goh,
  Sandhini Agarwal, Girish Sastry, Amanda Askell, Pamela Mishkin, Jack Clark,
  Gretchen Krueger, and Ilya Sutskever.
\newblock Learning transferable visual models from natural language
  supervision.
\newblock In Marina Meila and Tong Zhang, editors, {\em Proceedings of the 38th
  International Conference on Machine Learning, {ICML} 2021, 18-24 July 2021,
  Virtual Event}, volume 139 of {\em Proceedings of Machine Learning Research},
  pages 8748--8763. {PMLR}, 2021.

\bibitem{BM25}
Stephen~E. Robertson, Steve Walker, Susan Jones, Micheline Hancock{-}Beaulieu,
  and Mike Gatford.
\newblock Okapi at {TREC-3}.
\newblock In Donna~K. Harman, editor, {\em Proceedings of The Third Text
  REtrieval Conference, {TREC} 1994, Gaithersburg, Maryland, USA, November 2-4,
  1994}, volume 500-225 of {\em {NIST} Special Publication}, pages 109--126.
  National Institute of Standards and Technology {(NIST)}, 1994.

\bibitem{cc3m}
Piyush Sharma, Nan Ding, Sebastian Goodman, and Radu Soricut.
\newblock Conceptual captions: {A} cleaned, hypernymed, image alt-text dataset
  for automatic image captioning.
\newblock In Iryna Gurevych and Yusuke Miyao, editors, {\em Proceedings of the
  56th Annual Meeting of the Association for Computational Linguistics, {ACL}
  2018, Melbourne, Australia, July 15-20, 2018, Volume 1: Long Papers}, pages
  2556--2565. Association for Computational Linguistics, 2018.

\bibitem{lexmae}
Tao Shen, Xiubo Geng, Chongyang Tao, Can Xu, Xiaolong Huang, Binxing Jiao,
  Linjun Yang, and Daxin Jiang.
\newblock Lex{MAE}: Lexicon-bottlenecked pretraining for large-scale retrieval.
\newblock In {\em International Conference on Learning Representations}, 2023.

\bibitem{unified}
Tao Shen, Xiubo Geng, Chongyang Tao, Can Xu, Kai Zhang, and Daxin Jiang.
\newblock Unifier: {A} unified retriever for large-scale retrieval.
\newblock {\em CoRR}, abs/2205.11194, 2022.

\bibitem{sun-etal-2021-lightningdot}
Siqi Sun, Yen-Chun Chen, Linjie Li, Shuohang Wang, Yuwei Fang, and Jingjing
  Liu.
\newblock {L}ightning{DOT}: Pre-training visual-semantic embeddings for
  real-time image-text retrieval.
\newblock In {\em Proceedings of the 2021 Conference of the North American
  Chapter of the Association for Computational Linguistics: Human Language
  Technologies}, pages 982--997, Online, June 2021. Association for
  Computational Linguistics.

\bibitem{DBLP:journals/pami/WangLHL19}
Liwei Wang, Yin Li, Jing Huang, and Svetlana Lazebnik.
\newblock Learning two-branch neural networks for image-text matching tasks.
\newblock {\em {IEEE} Trans. Pattern Anal. Mach. Intell.}, 41(2):394--407,
  2019.

\bibitem{simlm}
Liang Wang, Nan Yang, Xiaolong Huang, Binxing Jiao, Linjun Yang, Daxin Jiang,
  Rangan Majumder, and Furu Wei.
\newblock Simlm: Pre-training with representation bottleneck for dense passage
  retrieval.
\newblock {\em CoRR}, abs/2207.02578, 2022.

\bibitem{cookie}
Keyu Wen, Jin Xia, Yuanyuan Huang, Linyang Li, Jiayan Xu, and Jie Shao.
\newblock {COOKIE:} contrastive cross-modal knowledge sharing pre-training for
  vision-language representation.
\newblock In {\em 2021 {IEEE/CVF} International Conference on Computer Vision,
  {ICCV} 2021, Montreal, QC, Canada, October 10-17, 2021}, pages 2188--2197.
  {IEEE}, 2021.

\bibitem{anserini}
Peilin Yang, Hui Fang, and Jimmy Lin.
\newblock Anserini: Enabling the use of lucene for information retrieval
  research.
\newblock In Noriko Kando, Tetsuya Sakai, Hideo Joho, Hang Li, Arjen~P. de
  Vries, and Ryen~W. White, editors, {\em Proceedings of the 40th International
  {ACM} {SIGIR} Conference on Research and Development in Information
  Retrieval, Shinjuku, Tokyo, Japan, August 7-11, 2017}, pages 1253--1256.
  {ACM}, 2017.

\bibitem{DBLP:conf/iclr/YangDSC18}
Zhilin Yang, Zihang Dai, Ruslan Salakhutdinov, and William~W. Cohen.
\newblock Breaking the softmax bottleneck: {A} high-rank {RNN} language model.
\newblock In {\em 6th International Conference on Learning Representations,
  {ICLR} 2018, Vancouver, BC, Canada, April 30 - May 3, 2018, Conference Track
  Proceedings}. OpenReview.net, 2018.

\end{thebibliography}
